\documentclass[a4paper,11pt]{esslli}
\usepackage{amsmath2000}
\usepackage{amstext2000}
\usepackage{amssymb}
\usepackage{stmaryrd}
\usepackage{chicago}
\usepackage{url}

\begin{document}

\newcommand{\CCSHANsquashleft}[2][1em]
    {\makebox[#1]{$\displaystyle #2 \hskip0pt minus1fill$}}

\newcommand{\CCSHANphrase}[1]{\emph{#1}}
\newcommand{\CCSHANdenote}[1]{\ensuremath{\llbracket}#1\ensuremath{\rrbracket}}
\newcommand{\CCSHANcompose}{\mathbin\circ}

\newcommand{\CCSHANmonad}{\mathord{\mathbb M}}
\newcommand{\CCSHANunit }{\eta}
\newcommand{\CCSHANbind }{\mathbin{\star}}
\newcommand{\CCSHANlift }{\ell}
\newcommand{\CCSHANeval }{\varepsilon}

\newcommand{\CCSHANapply  }{\ensuremath{A          }}
\newcommand{\CCSHANapplym }{\ensuremath{A_{\CCSHANmonad} }}
\newcommand{\CCSHANapplymp}{\ensuremath{A_{\CCSHANmonad}'}}

\def\CCSHANfun#1.#2{\mathop{\lambda\mathord{#1}.}#2}

\newcommand{\CCSHANjohn}{\textrm{John}}
\newcommand{\CCSHANmary}{\textrm{Mary}}
\newcommand{\CCSHANlike}{\textrm{like}}
\newcommand{\CCSHANhate}{\textrm{hate}}
\newcommand{\CCSHANsmoke}{\textrm{smoke}}

\newcommand{\CCSHANtyin }{:}
\newcommand{\CCSHANtyenv}{\rho}
\newcommand{\CCSHANtyans}{\omega}

\newcommand{\CCSHANshift}{\operatorname{shift}}
\newcommand{\CCSHANreset}{\operatorname{reset}}
\newcommand{\CCSHANid   }{\operatorname{id   }}

\newcommand{\CCSHANxlatev}[1]{\left\lfloor#1\right\rfloor}
\newcommand{\CCSHANxlatec}[1]{\left\lceil #1\right\rceil }
\newcommand{\CCSHANxlatee}[1]{\left\lceil #1\right\rceil }
\newcommand{\CCSHANxlated}[1]{     \llceil#1      \rrceil}
\newcommand{\CCSHANxlatef}[1]{     \llceil#1      \rrceil}
\newcommand{\CCSHANvoid  }{\mathord{!}}
\setcounter{page}{285}\setcounter{chapter}{23}
\chapter{Monads for natural language semantics}

\chapterauthor{Chung-chieh Shan}
\chapteraffiliation{Harvard University, 33 Oxford St, Cambridge, MA 02138, USA}
\chapteremail{ken@digitas.harvard.edu}

\begin{chapterabstract}
Accounts of semantic phenomena often involve extending types of meanings
and revising composition rules at the same time.  The concept of
\emph{monads} allows many such accounts---for intensionality, variable
binding, quantification and focus---to be stated uniformly and
compositionally.
\end{chapterabstract}

\section{Introduction}

The Montague grammar tradition formulates formal semantics for natural
languages in terms of the $\lambda$-calculus.  Each utterance is considered
a tree in which each leaf node is a lexical item whose meaning is a
(usually typed) value in the $\lambda$-calculus.  The leaf node meanings
then determine meanings for subtrees, through recursive application of one
or more \emph{composition rules}.  A composition rule specifies the meaning
of a tree in terms of the meanings of its sub-trees.  One simple
composition rule is function application:
\begin{equation}
\label{e:fa}
\CCSHANdenote{x\,y} = \CCSHANdenote{x}\bigl(\CCSHANdenote{y}\bigr) \CCSHANtyin \beta
\qquad\text{where $\CCSHANdenote{x}\CCSHANtyin\alpha\to\beta$ and $\CCSHANdenote{y}\CCSHANtyin\alpha$}.
\end{equation}
Here $\alpha$ and $\beta$ are type variables, and we denote function types
by $\to$.

To handle phenomena such as intensionality, variable binding,
quantification and focus, we often introduce new types in which to embed
existing aspects of meaning and accommodate additional ones.  Having
introduced new types, we then need to revise our composition rules to
reimplement existing functionality.  In this way, we often augment semantic
theories by simultaneously extending the types of meanings and stipulating
new composition rules.  When we augment a grammar, its original lexical
meanings and composition rules become invalid and require global renovation
(typically described as ``generalizing to the worst
case''~\cite{partee-development}).  Each time we consider a new aspect of
meaning, all lexical meanings and composition rules have to be revised.

Over the past decade, the category-theoretic concept of \emph{monads} has
gained popularity in computer science as a tool to structure denotational
semantics~\cite{moggi-abstract,moggi-notions} and functional
programs~\cite{wadler-comprehending,wadler-essence}.  When used to
structure computer programs, monads allow the substance of a computation to
be defined separately from the plumbing that supports its execution,
increasing modularity.  Many accounts of phenomena in natural language
semantics can also be phrased in terms of monads, thus clarifying the
account and simplifying the presentation.

In this paper, I will present the concept of monads and show how they can
be applied to natural language semantics.  To illustrate the approach, I
will use four monads to state analyses of well-known phenomena
\emph{uniformly} and \emph{compositionally}.  By ``uniformly'' I mean that,
even though the analyses make use of a variety of monads, they all invoke
monad primitives in the same way.  By ``compositionally'' I mean that the
analyses define composition rules in the spirit of Montague grammar.  After
presenting the monadic analyses, I will discuss combining monads to account
for interaction between semantic phenomena.

\section{Monadic analyses}

Intuitively, a monad is a transformation on types equipped with a
composition method for transformed values.  Formally, a monad is a triple
$(\CCSHANmonad, \CCSHANunit, \CCSHANbind)$, where $\CCSHANmonad$ is a type constructor (a map from
each type $\alpha$ to a corresponding type $\CCSHANmonad\alpha$), and $\CCSHANunit$ and
$\CCSHANbind$ are functions (pronounced ``unit'' and ``bind'')
\begin{equation}
\label{e:monad}
\CCSHANunit \CCSHANtyin \alpha \to \CCSHANmonad \alpha,\qquad
\CCSHANbind \CCSHANtyin \CCSHANmonad \alpha \to (\alpha \to \CCSHANmonad \beta) \to \CCSHANmonad \beta.
\end{equation}
These two functions are
polymorphic in the sense that they must be defined for all types $\alpha$
and $\beta$.  Roughly speaking, $\CCSHANunit$ specifies how ordinary values can
be injected into the monad, and $\CCSHANbind$ specifies how computations within
the monad compose with each other.\footnote{By definition, $\CCSHANunit$ and
$\CCSHANbind$ must satisfy left identity, right identity, and associativity:
\begin{subequations}
\label{e:monad-laws}
\begin{align}
\label{e:monad-law-li}
\CCSHANunit(a) \CCSHANbind k  &= k(a)
    &&\forall a\CCSHANtyin\alpha\text{, }k\CCSHANtyin\alpha\to\CCSHANmonad\beta,\\
\label{e:monad-law-ri}
m \CCSHANbind \CCSHANunit     &= m
    &&\forall m\CCSHANtyin\CCSHANmonad\alpha,\\
\label{e:monad-law-a}
\bigl(m\CCSHANbind k\bigr)\CCSHANbind l &= m\CCSHANbind\bigl(\CCSHANfun v.{k(v)\CCSHANbind l}\bigr)
    &&\forall m\CCSHANtyin\CCSHANmonad\alpha\text{, }
                k\CCSHANtyin\alpha\to\CCSHANmonad\beta\text{, }
                l\CCSHANtyin\beta\to\CCSHANmonad\gamma.
\end{align}
\end{subequations}\vspace{-\baselineskip}}
Some concrete examples follow.

\subsection{The powerset monad; interrogatives}

As a first example, consider sets.  Corresponding to each type $\alpha$ we
have a type $\alpha\to t$, the type of subsets of $\alpha$.  We
define\footnote{\label{f:ppp-strength}In this section and the next, we
treat types as sets in order to define the powerset and pointed powerset
monads.  These two monads do not exist in every model of the
$\lambda$-calculus.}
\begin{subequations}
\label{e:powerset-monad}
\begin{align}
\CCSHANmonad \alpha &= \alpha \to t
    &&\quad\forall \alpha,\\
\CCSHANunit (a)     &= \{a\} \CCSHANtyin \CCSHANmonad \alpha
    &&\quad\forall a\CCSHANtyin\alpha,\\
m \CCSHANbind k     &= \textstyle\bigcup_{a\in m} k(a) \CCSHANtyin \CCSHANmonad \beta
    &&\quad\forall m\CCSHANtyin\alpha\to t\text{, }k\CCSHANtyin\alpha\to\beta\to t.
\end{align}
\end{subequations}

The powerset monad is a crude model of non-determinism.  For example, the set of
individuals $m_1$ defined by
\begin{equation*}
m_1 = \{\CCSHANjohn,\CCSHANmary\} \CCSHANtyin \CCSHANmonad e
\end{equation*}
can be thought of as a non-deterministic individual---it is ambiguous
between John and Mary.  Similarly, the function $k_1$ defined by
\begin{equation*}
\begin{aligned}
k_1 &: e \to \CCSHANmonad (e\to t) \\
k_1(a) &= \{\CCSHANfun x.{\CCSHANlike(a,x)},\CCSHANfun x.{\CCSHANhate(a,x)}\} \CCSHANtyin \CCSHANmonad(e\to t)
\end{aligned}
\end{equation*}
maps each individual to a non-deterministic property.  To apply the
function $k_1$ to the individual $m_1$, we compute
\begin{equation*}
\begin{aligned}
m_1\CCSHANbind k_1 &= \textstyle\bigcup_{a\in\{\CCSHANjohn,\CCSHANmary\}}
           \{\CCSHANfun x.{\CCSHANlike(a,x)},\CCSHANfun x.{\CCSHANhate(a,x)}\}\\
         &= \{\CCSHANfun x.{\CCSHANlike(\CCSHANjohn,x)},\CCSHANfun x.{\CCSHANhate(\CCSHANjohn,x)},\\
         &\qquad\CCSHANfun x.{\CCSHANlike(\CCSHANmary,x)},\CCSHANfun x.{\CCSHANhate(\CCSHANmary,x)}\} \CCSHANtyin \CCSHANmonad(e\to t).
\end{aligned}
\end{equation*}
We see that the non-determinism in both $m_1$ and $k_1$ is carried through
to produce a $4$-way-ambiguous result.

Most words in natural language are not ambiguous in the way $m_1$ and $k_1$
are.  To upgrade an ordinary (deterministic) value of any type $\alpha$ to
the corresponding non-deterministic type $\CCSHANmonad \alpha$, we can apply
$\CCSHANunit$ to the ordinary value, say $\CCSHANjohn$:
\begin{equation}
\label{e:just-john}
\begin{aligned}
\CCSHANunit(\CCSHANjohn) &= \{\CCSHANjohn\} \CCSHANtyin \CCSHANmonad e,\\
\{\CCSHANjohn\}\CCSHANbind k_1
    &= \{\CCSHANfun x.{\CCSHANlike(\CCSHANjohn,x)},\CCSHANfun x.{\CCSHANhate(\CCSHANjohn,x)}\} \CCSHANtyin \CCSHANmonad(e\to t).
\end{aligned}
\end{equation}
Similarly, to convert an ordinary function to a non-deterministic function,
we can apply $\CCSHANunit$ to the output of the ordinary function, say $k_2$
below:
\begin{equation}
\label{e:just-like}
\begin{aligned}
k_2 &= \CCSHANfun a.{\CCSHANfun x.{\CCSHANlike(a,x)}} \CCSHANtyin e\to e\to t,\\
\CCSHANunit\CCSHANcompose k_2
    &= \CCSHANfun a.{\{\CCSHANfun x.{\CCSHANlike(a,x)}\}} \CCSHANtyin e\to\CCSHANmonad(e\to t),\\
m_1\CCSHANbind (\CCSHANunit\CCSHANcompose k_2)
    &= \{\CCSHANfun x.{\CCSHANlike(\CCSHANjohn,x)},\CCSHANfun x.{\CCSHANlike(\CCSHANmary,x)}\} \CCSHANtyin \CCSHANmonad(e\to t).
\end{aligned}
\end{equation}
In both~\eqref{e:just-john} and~\eqref{e:just-like}, an ordinary value is
made to work with a non-deterministic value by upgrading it to the
non-deterministic type.

Consider now the function application rule~\eqref{e:fa}.  We can regard it
as a two-argument function, denoted $\CCSHANapply$ and defined by
\begin{equation}
\label{e:a}
\begin{aligned}
\CCSHANapply &: (\alpha\to\beta)\to\alpha\to\beta,\\
\CCSHANapply(f)(x) &= f(x)\CCSHANtyin\beta \qquad \forall f\CCSHANtyin\alpha\to\beta\text{, }x\CCSHANtyin\alpha.
\end{aligned}
\end{equation}
We can lift ordinary function application $\CCSHANapply$ to non-deterministic
function application $\CCSHANapplym$, defined by
\begin{equation}
\label{e:am}
\begin{aligned}
\CCSHANapplym &: \CCSHANmonad(\alpha\to\beta)\to\CCSHANmonad\alpha\to\CCSHANmonad\beta,\\
\CCSHANapplym(f)(x) &= f \CCSHANbind \bigl[\CCSHANfun a. { x \CCSHANbind [\CCSHANfun b. {\CCSHANunit(a(b))}] }\bigr]
    \CCSHANtyin\CCSHANmonad\beta
    \\&\qquad\forall f\CCSHANtyin\CCSHANmonad(\alpha\to\beta)\text{, }x\CCSHANtyin\CCSHANmonad\alpha.
\end{aligned}
\end{equation}
Substituting~\eqref{e:powerset-monad} into~\eqref{e:am}, we get
\begin{equation}
\label{e:am-powerset-monad}
\CCSHANapplym(f)(x) = \{\, a(b) \mid a\in f \text{, } b\in x \,\} \subseteq \beta
    \qquad \forall f\subseteq\alpha\to\beta\text{, }x\subseteq\alpha.
\end{equation}

Just as the definition of~$\CCSHANapply$ in~\eqref{e:a} gives rise to the original
composition rule~\eqref{e:fa}, that is,
\begin{equation}
    \label{e:fa-a}
    \CCSHANdenote{x\,y} = \CCSHANapply(\CCSHANdenote{x})(\CCSHANdenote{y}),
\end{equation}
the definition of~$\CCSHANapplym$ in~\eqref{e:am} gives rise to the revised composition
rule
\begin{equation}
    \label{e:fa-am}
    \CCSHANdenote{x\,y} = \CCSHANapplym(\CCSHANdenote{x})(\CCSHANdenote{y}).
\end{equation}
For the powerset monad, this revised rule is the set-tolerant composition rule
in the alternative semantics analysis of interrogatives first proposed by
\citeN{hamblin-questions}.  In \shortciteANP{hamblin-questions}'s analysis,
the meaning of each interrogative constituent is a set of alternatives
available as answers to the question; this corresponds to the definition
of~$\CCSHANmonad$ in~\eqref{e:powerset-monad}.  By contrast, the meaning of each
non-interrogative constituent is a singleton set; this corresponds to the
definition of~$\CCSHANunit$ in~\eqref{e:powerset-monad}.

To support question-taking verbs (such as \CCSHANphrase{know} and \CCSHANphrase{ask}),
we (and~\shortciteANP{hamblin-questions}) need a secondary composition rule
in which $\CCSHANapply$ is lifted with respect to the function $f$ but not the
argument $x$:
\begin{equation}
\label{e:amp}
\begin{aligned}
\CCSHANdenote{x\,y} &= \CCSHANapplymp(\CCSHANdenote{x})(\CCSHANdenote{y}) \qquad \text{where}\\
\CCSHANapplymp &: \CCSHANmonad(\CCSHANmonad\alpha\to\beta)\to\CCSHANmonad\alpha\to\CCSHANmonad\beta,\\
\CCSHANapplymp(f)(x) &= f \CCSHANbind \bigl[\CCSHANfun a. { \CCSHANunit(a(x)) }\bigr]
    \CCSHANtyin\CCSHANmonad\beta
    \qquad \forall f\CCSHANtyin\CCSHANmonad(\CCSHANmonad\alpha\to\beta)\text{, }x\CCSHANtyin\CCSHANmonad\alpha.
\end{aligned}
\end{equation}
Substituting~\eqref{e:powerset-monad} into~\eqref{e:amp}, we get
\begin{equation}
\label{e:amp-powerset-monad}
\CCSHANapplymp(f)(x) = \{\, a(x) \mid a\in f \,\} \subseteq \beta
    \qquad \forall f\subseteq(\alpha\to t)\to\beta\text{, }x\subseteq\alpha,
\end{equation}
Note that, for any given pair of types of $\CCSHANdenote{x}$ and $\CCSHANdenote{y}$, at
most one of $\CCSHANapplym$~\eqref{e:am} and $\CCSHANapplymp$~\eqref{e:amp} can apply.
Thus the primary composition rule~\eqref{e:fa-am} and the secondary
composition rule~\eqref{e:amp} never conflict.

\subsection{The pointed powerset monad; focus}

A variation on the powerset monad~\eqref{e:powerset-monad} is the
\emph{pointed powerset monad}; it is implicitly involved in
\citeANP{rooth-focus}'s \citeyear{rooth-focus} account of focus in natural
language.  A \emph{pointed set} is a nonempty set with a distinguished
member.  In other words, a pointed set $x$ is a pair $x = (x_0,x_1)$, such
that $x_0$ is a member of the set $x_1$.  Define the pointed powerset monad
by
\begin{subequations}
\label{e:pointed-powerset-monad}
\begin{align}
\CCSHANmonad \alpha &= \bigl\{\, (x_0,x_1) \mid x_0\in x_1\subseteq\alpha \,\bigr\}
    &&\;\;\forall \alpha,\\
\CCSHANunit (a)     &= \bigl(a,\{a\}\bigr) \CCSHANtyin \CCSHANmonad \alpha
    &&\;\;\forall a\CCSHANtyin\alpha,\\
m \CCSHANbind k &= \textstyle
    \bigl( [k(m_0)]_0,
           \bigcup_{a\in m_1} [k(a)]_1 \bigr) \CCSHANtyin \CCSHANmonad \beta
    &&\;\;\forall m\CCSHANtyin\CCSHANmonad\alpha\text{, }k\CCSHANtyin\alpha\to\CCSHANmonad\beta.
\end{align}
\end{subequations}
This definition captures the intuition that we want to keep track of both a
set of non-deterministic alternatives and a particular alternative in the
set.  As with the powerset monad, the definition of $m\CCSHANbind k$ carries the
non-determinism in both $m$ and $k$ through to the result.

Substituting our new monad definition~\eqref{e:pointed-powerset-monad} into the
previously lifted application formula~\eqref{e:am} gives
\begin{equation}
\label{e:am-pointed-powerset-monad}
\begin{aligned}
\CCSHANapplym(f_0,f_1)(x_0,x_1)
    &= \bigl( f_0(x_0),
      \{\, a(b) \mid a\in f_1 \text{, } b\in x_1 \,\} \bigr) \CCSHANtyin\CCSHANmonad\beta\\
    &\qquad \forall (f_0,f_1)\CCSHANtyin\CCSHANmonad(\alpha\to\beta)\text{, }
                   (x_0,x_1)\CCSHANtyin\CCSHANmonad\alpha.
\end{aligned}
\end{equation}
This formula, in conjunction with the primary composition
rule~\eqref{e:fa-am}, is equivalent to \citeANP{rooth-focus}'s recursive
definition of focus semantic values.

Crucially, even though the pointed powerset monad extends our meaning types to
accommodate focus information, neither our definition of $\CCSHANapplym$
in~\eqref{e:am} nor our composition rule~\eqref{e:fa-am} needs to change from
before.  Moreover, the majority of our lexical meanings have nothing to do
with focus and thus need not change either.  For example, in the
hypothetical lexicon entry
$\CCSHANdenote{\text{\CCSHANphrase{John}}} = \CCSHANunit(\CCSHANjohn)$,
the upgrade from meaning type $e$ to meaning type $\CCSHANmonad e$ occurs
automatically due to the redefinition of $\CCSHANunit$.

\subsection{The reader monad; intensionality and variable binding}
\label{s:reader-monad}

Another monad often seen in computer science is the \emph{reader monad},
also known as the \emph{environment monad}.  This monad encodes
dependence of values on some given input.  To define the reader monad,
fix a type $\CCSHANtyenv$---say the type $s$ of possible worlds, or the
type $g$ of variable assignments---then let
\begin{subequations}
\label{e:reader-monad}
\begin{align}
\label{e:reader-monad-m}
\CCSHANmonad \alpha &= \CCSHANtyenv\to\alpha
    &&\quad\forall \alpha,\\
\label{e:reader-monad-u}
\CCSHANunit (a)     &= \CCSHANfun w.{a}\CCSHANtyin \CCSHANmonad \alpha
    &&\quad\forall a\CCSHANtyin\alpha,\\
\label{e:reader-monad-b}
m \CCSHANbind k     &= \CCSHANfun w.{k\bigl(m(w)\bigr)(w)} \CCSHANtyin \CCSHANmonad \beta
    &&\quad\forall m\CCSHANtyin\CCSHANmonad\alpha\text{, }k\CCSHANtyin\alpha\to\CCSHANmonad\beta.
\end{align}
\end{subequations}
Note how the definition of $m\CCSHANbind k$ threads the input $w$ through both
$m$ and $k$ to produce the result.  To see this threading process in
action, let us once again substitute our monad
definition~\eqref{e:reader-monad} into the definition of $\CCSHANapplym$
in~\eqref{e:am}:
\begin{equation}
\label{e:am-reader-monad}
\begin{aligned}
\CCSHANapplym(f)(x)
    &= \CCSHANfun w.f(w)\bigl(x(w)\bigr) \CCSHANtyin\CCSHANmonad\beta
    \qquad \forall f\CCSHANtyin\CCSHANmonad(\alpha\to\beta)\text{, } x\CCSHANtyin\CCSHANmonad\alpha.
\end{aligned}
\end{equation}
For $\CCSHANtyenv=s$, we can think of $\CCSHANmonad$ as the intensionality monad,
noting that~\eqref{e:am-reader-monad} is exactly the usual extensional
composition rule.  While words such as \CCSHANphrase{student} and \CCSHANphrase{know}
have meanings that depend on the possible world $w$, words such as
\CCSHANphrase{is} and \CCSHANphrase{and} do not.  We can upgrade the latter
by applying $\CCSHANunit$.

For $\CCSHANtyenv=g$, we can think of $\CCSHANmonad$ as the variable binding monad,
noting that~\eqref{e:am-reader-monad} is the usual assignment-preserving
composition rule.  Except for pronominals, most word meanings do not refer
to the variable assignment.  Thus we can upgrade the majority of word
meanings by applying $\CCSHANunit$.

If we substitute the same monad definition~\eqref{e:reader-monad} into the
secondary composition rule~\eqref{e:amp}, the result is
\begin{equation}
\label{e:amp-reader-monad}
\begin{aligned}
\CCSHANapplymp(f)(x)
    &= \CCSHANfun w.f(w)(x) \CCSHANtyin\CCSHANmonad\beta
    \qquad \forall f\CCSHANtyin\CCSHANmonad(\CCSHANmonad\alpha\to\beta)\text{, }
                   x\CCSHANtyin\CCSHANmonad\alpha.
\end{aligned}
\end{equation}
For $\CCSHANtyenv=s$, this is the intensional composition rule; it handles
sentence-taking verbs such as \emph{know} and \emph{believe} (of type
$s\to(s\to t)\to e\to t$) by allowing them to take arguments of type $s\to
t$ rather than type $t$.  The monad laws, by the way, guarantee that
$\CCSHANapplymp(f)(x) = \CCSHANapplym(f)\bigl(\CCSHANunit(x)\bigr)$ for all $f$ and $x$; the
function $\CCSHANunit$ (in this case a map from $s\to t$ to $s\to s\to t$) is
simply the intension (up) operator, usually written~$^\wedge$.

For $\CCSHANtyenv=g$, the same formula~\eqref{e:amp-reader-monad} is often
involved in accounts of quantification that assume quantifier raising at
LF, such as that in \citeN{heim-semantics}.  It handles raised quantifiers
(of type $g\to(g\to t)\to t$) by allowing them to take arguments of type
$g\to t$ rather than type $t$.  The function $\CCSHANunit$ (in this case a map
from $g\to t$ to $g\to g\to t$) is simply the variable abstraction
operator.

\subsection{The continuation monad; quantification}

\citeN{barker-continuations} proposed an analysis of quantification in
terms of \emph{continuations}.  The basic idea is to \emph{continuize} a
grammar by replacing each meaning type $\alpha$ with its corresponding
continuized type $(\alpha\to t)\to t$ throughout.  As a special case, the
meaning type of NPs is changed from $e$ to $(e\to t)\to t$, matching the
original treatment of English quantification by~\citeN{montague-proper}.

In general, for any fixed type $\CCSHANtyans$ (say $t$), we can define a
\emph{continuation monad} with \emph{answer type} $\CCSHANtyans$:
\begin{subequations}
\label{e:cont-monad}
\begin{align}
\CCSHANmonad \alpha &= (\alpha\to\CCSHANtyans)\to\CCSHANtyans
    &&\quad\forall \alpha,\\
\CCSHANunit (a)     &= \CCSHANfun c.{c(a)}\CCSHANtyin \CCSHANmonad \alpha
    &&\quad\forall a\CCSHANtyin\alpha,\\
m \CCSHANbind k     &= \CCSHANfun c.{m\bigl(\CCSHANfun a.{k(a)(c)}\bigr)} \CCSHANtyin \CCSHANmonad \beta
    &&\quad\forall m\CCSHANtyin\CCSHANmonad\alpha\text{, }k\CCSHANtyin\alpha\to\CCSHANmonad\beta.
\end{align}
\end{subequations}
The value $c$ manipulated in these definitions is known as the
\emph{continuation}.  Intuitively, ``the continuation represents an entire
(default) future for the computation''~\cite{kelsey-r5rs}.  Each value of type
$\CCSHANmonad\alpha$ must turn a continuation (of type $\alpha\to\CCSHANtyans$) into an answer
(of type $\CCSHANtyans$).  The most obvious way to do so, encoded in the definition of
$\CCSHANunit$ above, is to feed the continuation a value of type $\alpha$:
\begin{subequations}
\begin{align}
\CCSHANdenote{\text{\CCSHANphrase{John}}} = \CCSHANunit(\CCSHANjohn)
    &= \CCSHANfun c.{c(\CCSHANjohn)} \CCSHANtyin \CCSHANmonad e,\\
\CCSHANdenote{\text{\CCSHANphrase{smokes}}} = \CCSHANunit(\CCSHANsmoke)
    &= \CCSHANfun c.{c(\CCSHANsmoke)} \CCSHANtyin \CCSHANmonad (e\to t).
\end{align}
\end{subequations}
To compute the meaning of \CCSHANphrase{John smokes}, we first substitute
our monad definition~\eqref{e:cont-monad} into the primary composition
operation $\CCSHANapplym$~\eqref{e:am}:
\begin{equation}
\label{e:am-cont-monad}
\begin{aligned}
\CCSHANapplym(f)(x) &= \CCSHANfun c.{f\bigl(\CCSHANfun g.{x\bigl(\CCSHANfun y.{c(g(y))}\bigr)}\bigr)} \CCSHANtyin\CCSHANmonad\beta
    \\&\qquad \forall f\CCSHANtyin\CCSHANmonad(\alpha\to\beta)\text{, } x\CCSHANtyin\CCSHANmonad\alpha.
\end{aligned}
\end{equation}
Letting $f=\CCSHANunit(\CCSHANsmoke)$ and $x=\CCSHANunit(\CCSHANjohn)$ then gives
\begin{align*}
\CCSHANsquashleft{\CCSHANdenote{\text{\CCSHANphrase{John smokes}}}
  = \CCSHANfun c.{\CCSHANunit(\CCSHANsmoke)\bigl(\CCSHANfun g.{\CCSHANunit(\CCSHANjohn)\bigl(\CCSHANfun y.{c(g(y))}\bigr)}\bigr)}} \\
& = \CCSHANfun c.{\CCSHANunit(\CCSHANjohn)\bigl(\CCSHANfun y.{c(\CCSHANsmoke(y))}\bigr)}
  = \CCSHANfun c.{c(\CCSHANsmoke(\CCSHANjohn))} \CCSHANtyin \CCSHANmonad t.
\end{align*}
In the second step above, note how the term $\CCSHANfun
y.{c(\CCSHANsmoke(y))}$ represents the future for the computation of
\CCSHANdenote{\CCSHANphrase{John}}, namely to check whether he smokes, then
pass the result to the context $c$ containing the clause.  If
\CCSHANphrase{John smokes} is the main clause, then
the context $c$ is simply the identity function $\CCSHANid_\CCSHANtyans$.
We define an evaluation operator $\CCSHANeval \CCSHANtyin \CCSHANmonad
\CCSHANtyans\to\CCSHANtyans$ by $\CCSHANeval(m) =
m(\CCSHANid_\CCSHANtyans)$.  Fixing $\CCSHANtyans = t$, we then have
$\CCSHANeval\bigl(\CCSHANdenote{\text{\CCSHANphrase{John smokes}}}\bigr) =
\CCSHANsmoke(\CCSHANjohn)$, as desired.

Continuing to fix $\CCSHANtyans = t$, we can specify a meaning for
\CCSHANphrase{everyone}:
\begin{equation}
\label{e:everyone}
\CCSHANdenote{\text{\CCSHANphrase{everyone}}}
    = \CCSHANfun c.{\mathop{\forall x.} c(x)} \CCSHANtyin \CCSHANmonad e.
\end{equation}
This formula is not of the form $\CCSHANfun c.{c(\dots)}$.  In other words,
the meaning of \CCSHANphrase{everyone} non-trivially manipulates the
continuation, and so cannot be obtained from applying $\CCSHANunit$ to an
ordinary value.  Using the continuized composition
rule~\eqref{e:am-cont-monad}, we now compute a denotation for
\CCSHANphrase{everyone smokes}:
\begin{align*}
\CCSHANsquashleft{\CCSHANdenote{\text{\CCSHANphrase{everyone smokes}}}
  = \CCSHANfun c.{\CCSHANunit(\CCSHANsmoke)\bigl(\CCSHANfun g.{\CCSHANdenote{\text{\CCSHANphrase{everyone}}}\bigl(\CCSHANfun y.{c(g(y))}\bigr)}\bigr)}} \\
& = \CCSHANfun c.{\CCSHANdenote{\text{\CCSHANphrase{everyone}}}\bigl(\CCSHANfun y.{c(\CCSHANsmoke(y))}\bigr)}
  = \CCSHANfun c.{\mathop{\forall x.} c(\CCSHANsmoke(x))} \CCSHANtyin \CCSHANmonad t,
\end{align*}
giving $\CCSHANeval\bigl(\CCSHANdenote{\text{\CCSHANphrase{everyone
smokes}}}\bigr) = \mathop{\forall x.} \CCSHANsmoke(x) \CCSHANtyin t$,
as desired.

The main theoretical advantage of this analysis is that it is a
compositional, in-situ analysis that does not invoke quantifier raising.
Moreover, note that a grammar continuized is still a grammar---the
continuized composition rule~\eqref{e:am-cont-monad} is perfectly
interpretable using the standard machinery of Montague grammar.  In
particular, we do not invoke any type ambiguity or flexibility as proposed
by \citeN{partee-generalized} and \citeN{hendriks-studied}; the
interpretation mechanism performs no type-shifting at ``run-time''.

This desirable property also holds of the other monadic analyses I have
presented.  For instance, in a grammar with intensionality, meanings that
use intensionality (for example \CCSHANdenote{\CCSHANphrase{student}}) are identical in
type to meanings that do not (for example \CCSHANdenote{\CCSHANphrase{is}}).  The
interpretation mechanism does not dynamically shift the type of \CCSHANphrase{is}
to match that of \CCSHANphrase{student}.

It is worth relating the present analysis to the computer science
literature.  \citeN{danvy-abstracting} studied \emph{composable
continuations}, which they manipulated using two operators ``$\CCSHANshift$'' and
``$\CCSHANreset$''.  We can define $\CCSHANshift$ and $\CCSHANreset$ for the continuation
monad by~\cite{wadler-composable}
\begin{alignat}{2}
\CCSHANshift&= \CCSHANfun h.{\CCSHANfun c.{\CCSHANeval\bigl(h(\CCSHANfun a.{\CCSHANfun c'.{c'(c(a))}})\bigr)}}
          && : \bigl((\alpha \to \CCSHANmonad\CCSHANtyans) \to \CCSHANmonad\CCSHANtyans\bigr) \to \CCSHANmonad\alpha,\\
\CCSHANreset&= \CCSHANfun m.{\CCSHANfun c.{c\bigl(\CCSHANeval(m)\bigr)}}
          && : \CCSHANmonad\CCSHANtyans \to \CCSHANmonad\CCSHANtyans.
\end{alignat}
Assuming that the $\forall$ operator is of type $(e\to t)\to t$, the
meaning of \CCSHANphrase{everyone} specified in~\eqref{e:everyone} is
simply $\CCSHANshift\bigl(\CCSHANfun
c.{(\CCSHANunit\CCSHANcompose\forall)(\CCSHANeval\CCSHANcompose c)}\bigr)$.
To encode scope islands, \citeANP{barker-continuations} implicitly used
$\CCSHANreset$.
\citeN{filinski-layered} proved that, in a certain sense, composable
continuations can simulate monads.

\section{Combining monads}

Having placed various semantic phenomena in a monadic framework, we now ask
a natural question: Can we somehow combine monads in a modular fashion to
characterize interaction between semantic phenomena, for example between
intensionality and quantification?

Unfortunately, there exists no general construction for composing two
arbitrary monads, say $(\CCSHANmonad_1,\CCSHANunit_1,\CCSHANbind_1)$ and
$(\CCSHANmonad_2,\CCSHANunit_2,\CCSHANbind_2)$, into a new monad of the form
$(\CCSHANmonad_3 = \CCSHANmonad_1\CCSHANcompose\CCSHANmonad_2,\CCSHANunit_3,\CCSHANbind_3)$
\cite{king-combining,jones-composing}.
One might still hope to specialize and combine monads with additional
structure, to generalize and combine monads as instances of a broader
concept, or even to find that obstacles in combining monads are reflected
in semantic constraints in natural language.

Researchers in denotational semantics of programming languages have made
several proposals towards combining monadic functionality, none of which
are completely satisfactory
\cite{moggi-abstract,steele-building,liang-interpreter,espinosa-semantic,filinski-layered}.
In this section, I will relate one prominent approach to natural language
semantics.

\subsection{Monad morphisms}
\label{s:monad-morphisms}

One approach to combining monads, taken by \shortciteANP{moggi-abstract},
\shortciteANP{liang-interpreter}, and \shortciteANP{filinski-layered}, is to
compose \emph{monad morphisms} instead of monads themselves.  A monad
morphism (also known as a \emph{monad transformer} or a \emph{monad
layering}) is a map from monads to monads; it takes an arbitrary monad and
transforms it into a new monad, presumably defined in terms of the old
monad and supporting a new layer of functionality.  For instance, given any
monad $(\CCSHANmonad_1, \CCSHANunit_1, \CCSHANbind_1)$ and fixing a type $\CCSHANtyenv$, the
\emph{reader monad morphism} constructs a new monad $(\CCSHANmonad_2, \CCSHANunit_2,
\CCSHANbind_2)$, defined by
\begin{subequations}
\label{e:reader-monadx}
\begin{align}
\label{e:reader-monadx-m}
\CCSHANmonad_2 \alpha &= \CCSHANtyenv\to\CCSHANmonad_1\alpha
    &&\quad\forall \alpha,\\
\label{e:reader-monadx-u}
\CCSHANunit_2 (a)     &= \CCSHANfun w.{\CCSHANunit_1(a)}\CCSHANtyin \CCSHANmonad_2 \alpha
    &&\quad\forall a\CCSHANtyin\alpha,\\
\label{e:reader-monadx-b}
m \CCSHANbind_2 k     &= \CCSHANfun w.{\bigl[m(w) \CCSHANbind_1 \CCSHANfun a.{k(a)(w)}\bigr]} \CCSHANtyin \CCSHANmonad_2 \beta
   &&\quad\forall m\CCSHANtyin\CCSHANmonad_2\alpha\text{,}\\[-3pt]
   &&&\quad\hphantom{\forall}k\CCSHANtyin\alpha\to\CCSHANmonad_2\beta.\notag
\end{align}
\end{subequations}
If we let the old monad $(\CCSHANmonad_1, \CCSHANunit_1, \CCSHANbind_1)$ be the
\emph{identity monad}, defined by $\CCSHANmonad_1\alpha = \alpha$, $\CCSHANunit_1(a) =
a$, and $m\CCSHANbind_1 k = k(m)$, then the new monad~\eqref{e:reader-monadx} is
just the reader monad~\eqref{e:reader-monad}.  If we let the old monad be
some other monad---even the reader monad itself---the new monad adds reader
functionality.

By definition, each monad morphism must specify how to embed computations
inside the old monad into the new monad.  More precisely, each monad
morphism must provide a function (pronounced ``lift'')
\begin{equation}
\label{e:lift}
    \CCSHANlift : \CCSHANmonad_1\alpha \to \CCSHANmonad_2\alpha,
\end{equation}
polymorphic in $\alpha$.\footnote{By definition,
$\CCSHANlift$ must satisfy naturality:
\begin{subequations}
\begin{align}
\CCSHANlift\bigl(\CCSHANunit_1(a)\bigr) &= \CCSHANunit_2(a)
    &&\forall a\CCSHANtyin\alpha,\\
\CCSHANlift\bigl(m\CCSHANbind_1 k\bigr) &= \CCSHANlift(m) \CCSHANbind_2 (\CCSHANlift\CCSHANcompose k)
    &&\forall m\CCSHANtyin\CCSHANmonad_1\alpha\text{, }
              k\CCSHANtyin\alpha\to\CCSHANmonad_1\beta.
\end{align}
\end{subequations}\vspace{-\baselineskip}}
For the reader monad morphism, $\CCSHANlift$ is defined by
\begin{equation}
\begin{aligned}
\CCSHANlift(m) &= \CCSHANfun w.{m}\CCSHANtyin\CCSHANmonad_2\alpha
    &&\quad\forall m\CCSHANtyin\CCSHANmonad_1\alpha.
\end{aligned}
\end{equation}

The continuation monad also generalizes to a monad morphism.  Fixing an
answer type $\CCSHANtyans$, the \emph{continuation monad morphism} takes any
monad $(\CCSHANmonad_1, \CCSHANunit_1, \CCSHANbind_1)$ to the monad $(\CCSHANmonad_2, \CCSHANunit_2,
\CCSHANbind_2)$ defined by
\begin{subequations}
\label{e:cont-monadx}
\begin{align}
\label{e:cont-monadx-m}
\CCSHANmonad_2 \alpha &= (\alpha\to\CCSHANmonad_1\CCSHANtyans)\to\CCSHANmonad_1\CCSHANtyans
    &&\quad\forall \alpha,\\
\label{e:cont-monadx-u}
\CCSHANunit_2 (a)     &= \CCSHANfun c.{c(a)}\CCSHANtyin \CCSHANmonad_2 \alpha
    &&\quad\forall a\CCSHANtyin\alpha,\\
\label{e:cont-monadx-b}
m \CCSHANbind_2 k     &= \CCSHANfun c.{m\bigl(\CCSHANfun a.{k(a)(c)}\bigr)} \CCSHANtyin \CCSHANmonad_2 \beta
    &&\quad\forall m\CCSHANtyin\CCSHANmonad_2\alpha\text{, }k\CCSHANtyin\alpha\to\CCSHANmonad_2\beta.
\end{align}
\end{subequations}
The lifting function $\CCSHANlift$ for the continuation monad morphism is defined by
\begin{equation}
\begin{aligned}
\CCSHANlift(m) &= \CCSHANfun c.{(m\CCSHANbind_1 c)}\CCSHANtyin\CCSHANmonad_2\alpha
    &&\quad\forall m\CCSHANtyin\CCSHANmonad_1\alpha.
\end{aligned}
\end{equation}

Monad morphisms can be freely composed with each other, though the order of
composition is significant.  Applying the continuation monad morphism to
the reader monad is equivalent to applying to the identity monad the
composition of the continuation monad morphism and the reader monad
morphism, and yields a monad with type constructor $\CCSHANmonad\alpha =
(\alpha\to\CCSHANtyenv\to\CCSHANtyans)\to\CCSHANtyenv\to\CCSHANtyans$.  Applying the reader monad
morphism to the continuation monad is equivalent to applying to the
identity monad the composition of the reader monad morphism and the
continuation monad morphism, and yields a different monad, with type
constructor $\CCSHANmonad\alpha = \CCSHANtyenv\to(\alpha\to\CCSHANtyans)\to\CCSHANtyans$.

\subsection{Translating monads to monad morphisms}
\label{s:translating-monads}

The monad morphisms~\eqref{e:reader-monadx} and~\eqref{e:cont-monadx}
may appear mysterious, but we can in fact obtain them from their monad
counterparts~\eqref{e:reader-monad} and~\eqref{e:cont-monad} via a
mechanical translation.  The translation takes a monad
$(\CCSHANmonad_0,\CCSHANunit_0,\CCSHANbind_0)$ whose $\CCSHANunit_0$ and
$\CCSHANbind_0$ operations are $\lambda$-terms, and produces a
morphism mapping any old monad
$(\CCSHANmonad_1,\CCSHANunit_1,\CCSHANbind_1)$ to a new monad
$(\CCSHANmonad_2,\CCSHANunit_2,\CCSHANbind_2)$.  The translation is defined
recursively on the structure of $\lambda$-types and $\lambda$-terms, as
follows.\footnote{Among other things, the translation requires
$\CCSHANmonad_0$ to be defined as a $\lambda$-type, and $\CCSHANunit_0$ and
$\CCSHANbind_0$ to be defined as $\lambda$-terms.  Thus the translation
cannot apply to the powerset and pointed powerset monads (see
footnote~\ref{f:ppp-strength}).  Nevertheless, any monad morphism
(including ones produced by the translation) can be applied to any monad
(including these two monads).}

Every type $\tau$ is either a function type or a base type.  A function
type has the form $\tau_1\to\tau_2$, where $\tau_1$ and $\tau_2$ are types.
A base type $\iota$ is a type fixed in $\CCSHANmonad_0$ ($\CCSHANtyenv$ and
$\CCSHANtyans$ in our cases), a polymorphic type variable ($\alpha$ and
$\beta$ as appearing in $\CCSHANunit$ and
$\CCSHANbind$~\eqref{e:monad} and $\CCSHANlift$~\eqref{e:lift}), or the
terminal type $\CCSHANvoid$ (also known as the unit type or the void type).
For every type $\tau$, we recursively define its \emph{computation
translation} $\CCSHANxlatec{\tau}$ and its \emph{value translation}
$\CCSHANxlatev{\tau}$:
\begin{alignat}{3}
\label{e:xlate-type}
\CCSHANxlatec{\iota} &= \CCSHANmonad_1\iota, &\qquad
\CCSHANxlatev{\iota} &=         \iota, &\qquad
\CCSHANxlatec{\tau_1\to\tau_2} = \CCSHANxlatev{\tau_1\to\tau_2}
    &= \CCSHANxlatev{\tau_1}\to\CCSHANxlatec{\tau_2},
\end{alignat}
where $\iota$ is any base type.

Each term $e:\tau$ is an application term, an abstraction term, a variable
term, or the terminal term.  An application term has the form
${(e_1\CCSHANtyin\tau_1\to\tau_2)}\mathbin{}{(e_2\CCSHANtyin\tau_1)}\CCSHANtyin\tau_2$, where
$e_1$ and $e_2$ are terms.  An abstraction term has the form
$(\CCSHANfun x\CCSHANtyin\tau_1.{e\CCSHANtyin\tau_2})\CCSHANtyin\tau_1\to\tau_2$, where $e$ is a term.
A variable term has the form $x\CCSHANtyin\tau$, where $x$ is the name of a
variable of type $\tau$.  The terminal term is $\CCSHANvoid\CCSHANtyin\CCSHANvoid$ and
represents the unique value of the terminal type $\CCSHANvoid$.  For every term
$e\CCSHANtyin\tau$, we recursively define its \emph{term translation}
$\CCSHANxlatee{e}\CCSHANtyin\CCSHANxlatec{\tau}$:
\begin{subequations}
\label{e:xlate-term}
\begin{alignat}{2}
\CCSHANxlatee{\bigl.(e_1\CCSHANtyin\iota\to\tau_1\to\dotsb\to\tau_n\to\iota')
              (e_2\CCSHANtyin\iota)\bigr.}
  & = \notag\\
\label{e:xlate-term-ab}
  && \makebox[4em]{\hskip4em minus1fill$\displaystyle
     \CCSHANfun y_1\CCSHANtyin\CCSHANxlatev{\tau_1}.{ \dotso
     \CCSHANfun y_n\CCSHANtyin\CCSHANxlatev{\tau_n}.{ \bigl[ \CCSHANxlatee{e_2} \CCSHANbind_1
       \bigl(\CCSHANfun y_0\CCSHANtyin\iota. {\CCSHANxlatee{e_1}(y_0)\dotso(y_n)}\bigr) \bigr] }},
  $}\displaybreak[0]\\
\label{e:xlate-term-af}
\CCSHANxlatee{\bigl.(e_1\CCSHANtyin(\tau_1\to\tau_2)\to\tau_3)
              (e_2\CCSHANtyin\tau_1\to\tau_2)\bigr.}
  & = \CCSHANxlatee{e_1} \bigl(\CCSHANxlatee{e_2}\bigr),\displaybreak[1]\\
\label{e:xlate-term-l}
\CCSHANxlatee{\CCSHANfun x\CCSHANtyin\tau.{e}}
  & = \CCSHANfun x\CCSHANtyin\CCSHANxlatev{\tau}.\CCSHANxlatee{e},\displaybreak[1]\\
\label{e:xlate-term-vb}
\CCSHANxlatee{x\CCSHANtyin\iota}
  & = \CCSHANunit_1(x),\displaybreak[0]\\
\label{e:xlate-term-vf}
\CCSHANxlatee{x\CCSHANtyin\tau_1\to\tau_2}
  & = x,\displaybreak[0]\\
\label{e:xlate-term-void}
\CCSHANxlatee{\CCSHANvoid}
  & = \CCSHANunit_1(\CCSHANvoid),
\end{alignat}
\end{subequations}
where $\iota$ and $\iota'$ are any base types, and $y_0,\dotsc,y_n$ are
fresh variable names.

Finally, to construct the new monad
$(\CCSHANmonad_2,\CCSHANunit_2,\CCSHANbind_2)$, we specify
\begin{equation}
\label{e:xlate-monad}
\begin{gathered}
\CCSHANmonad_2\alpha = \CCSHANxlatec{\CCSHANmonad_0\alpha}, \qquad
\CCSHANunit_2        = \CCSHANxlatee{\CCSHANunit_0}, \qquad
\CCSHANbind_2        = \CCSHANxlatee{\CCSHANbind_0}, \\
\CCSHANlift(m)
    = \CCSHANxlatee{\CCSHANfun f\CCSHANtyin\CCSHANvoid\to\alpha.{\,\bigl.\CCSHANunit_0(f(\CCSHANvoid))\bigr.}}(\CCSHANfun \CCSHANvoid\CCSHANtyin\CCSHANvoid.{m})
    \quad \forall m\CCSHANtyin\CCSHANmonad_1\alpha.
\end{gathered}
\end{equation}

To illustrate this translation, let us expand out
$\CCSHANmonad_2$ and $\CCSHANbind_2$ in the special
case where $(\CCSHANmonad_0,\CCSHANunit_0,\CCSHANbind_0)$ is the reader
monad~\eqref{e:reader-monad}.  From the type translation
rules~\eqref{e:xlate-type} and the specification of
$\CCSHANmonad_2$ in~\eqref{e:xlate-monad}, we have
\begin{equation*}
\CCSHANmonad_2\alpha = \CCSHANxlatec{\CCSHANtyenv\to\alpha}
               = \CCSHANxlatev{\CCSHANtyenv}\to\CCSHANxlatec{\alpha}
               = \CCSHANtyenv\to\CCSHANmonad_1\alpha,
\end{equation*}
matching~\eqref{e:reader-monadx-m} as desired.  From the term translation
rules~\eqref{e:xlate-term} and the specification of
$\CCSHANbind_2$ in~\eqref{e:xlate-monad}, we have
\begin{alignat*}{2}
\CCSHANbind_2
    &= \CCSHANxlatee{
       \CCSHANfun m\CCSHANtyin\CCSHANtyenv\to\alpha.{
       \CCSHANfun k\CCSHANtyin\alpha\to\CCSHANtyenv\to\beta.{
       \CCSHANfun w\CCSHANtyin\CCSHANtyenv.{
           k\bigl(m(w)\bigr)(w)
       }}}}
    && \text{ by \eqref{e:reader-monad-b}} \\
    &= \CCSHANfun m\CCSHANtyin\CCSHANtyenv\to\CCSHANmonad_1\alpha.{
       \CCSHANfun k\CCSHANtyin\alpha\to\CCSHANtyenv\to\CCSHANmonad_1\beta.{
       \CCSHANfun w\CCSHANtyin\CCSHANtyenv.{
           \CCSHANxlatee{k\bigl(m(w)\bigr)(w)}
       }}}
    && \text{ by \eqref{e:xlate-term-l}},
\end{alignat*}
in which
\begin{alignat*}{2}
\CCSHANsquashleft{\CCSHANxlatee{k\bigl(m(w)\bigr)(w)}
     = \CCSHANunit_1(w) \CCSHANbind_1 \CCSHANfun y_0\CCSHANtyin\CCSHANtyenv. \CCSHANxlatee{k\bigl(m(w)\bigr)}(y_0)}&
    && \text{ by \eqref{e:xlate-term-ab}, \eqref{e:xlate-term-vb}} \\
    &= \CCSHANxlatee{k\bigl(m(w)\bigr)}(w)
    && \text{ by \eqref{e:monad-law-li}} \\
    &= \bigl(\CCSHANxlatee{w} \CCSHANbind_1 \CCSHANfun y_0\CCSHANtyin\CCSHANtyenv.{\CCSHANxlatee{m}(y_0)}\bigr)
       \CCSHANbind_1 \CCSHANfun y_0\CCSHANtyin\alpha.{\CCSHANxlatee{k}(y_0)(w)}
    && \text{ by \eqref{e:xlate-term-ab}} \\
    &= \bigl(\CCSHANunit_1(w) \CCSHANbind_1 \CCSHANfun y_0\CCSHANtyin\CCSHANtyenv.{m(y_0)}\bigr)
       \CCSHANbind_1 \CCSHANfun y_0\CCSHANtyin\alpha.{k(y_0)(w)}
    && \text{ by \eqref{e:xlate-term-vb}, \eqref{e:xlate-term-vf}} \\
    &= m(w) \CCSHANbind_1 \CCSHANfun y_0\CCSHANtyin\alpha.{k(y_0)(w)}
    && \text{ by \eqref{e:monad-law-li}},
\end{alignat*}
matching~\eqref{e:reader-monadx-b} as desired.

The intuition behind our translation is to treat the $\lambda$-calculus
with which $(\CCSHANmonad_0,\CCSHANunit_0,\CCSHANbind_0)$ is defined as a
programming language whose terms may have computational side effects.  Our
translation specifies a semantics for this programming language in terms of
$(\CCSHANmonad_1,\CCSHANunit_1,\CCSHANbind_1)$ that is call-by-value and
that allows side effects only at base types.  That the semantics is
call-by-value rather than call-by-name is reflected in the type translation
rules~\eqref{e:xlate-type}, where we define
$\CCSHANxlatev{\tau_1\to\tau_2}$ to be
$\CCSHANxlatev{\tau_1}\to\CCSHANxlatec{\tau_2}$ rather than
$\CCSHANxlatec{\tau_1}\to\CCSHANxlatec{\tau_2}$.  That side effects occur
only at base types is also reflected in the rules, where we define
$\CCSHANxlatec{\tau_1\to\tau_2}$ to be $\CCSHANxlatev{\tau_1\to\tau_2}$
rather than $\CCSHANmonad_1\CCSHANxlatev{\tau_1\to\tau_2}$.  Overall, our
translation is a hybrid between the call-by-name Algol
translation~\cite[\S3.1.2]{benton-appsem} and the standard call-by-value
translation~(\citeNP[\S8]{wadler-comprehending};
\shortciteANP[\S3.1.3]{benton-appsem}).

\subsection{A call-by-name translation of monads}

Curiously, the semantic types generated by monad morphisms seem sometimes
not powerful enough.
As noted at the end of~\S\ref{s:monad-morphisms}, the reader and
continuation monad morphisms together give rise to two different monads,
depending on the order in which we compose the monad morphisms.  Fixing
$\CCSHANtyenv = s$ for the reader monad morphism and $\CCSHANtyans = t$ for the
continuation monad morphism, the two combined monads have the type
constructors
\begin{align}
\label{e:combined-m}
\CCSHANmonad_{cr}\alpha &= (\alpha\to s\to t)\to s\to t,&
\CCSHANmonad_{rc}\alpha &= s\to(\alpha\to t)\to t.
\end{align}
Consider now sentences such as
\begin{equation}
\text{
John wanted to date every professor (at the party).
}
\end{equation}
This sentence has a reading where every professor at the party is a
person John wanted to date, but John may not be aware that they are
professors.  On this reading, note that the world where the property of
professorship is evaluated is distinct from the world where the property of
dating is evaluated.  Therefore, assuming that \CCSHANphrase{to date every
professor} is a constituent, its semantic type should mention $s$ in
contravariant position at least twice.  Unfortunately, the type
constructors $\CCSHANmonad_{cr}$ and
$\CCSHANmonad_{rc}$~\eqref{e:combined-m} each have only one such
occurrence, so neither $\CCSHANmonad_{cr}t$ nor $\CCSHANmonad_{rc}t$ can be
the correct type.

Intuitively, the semantic type of \CCSHANphrase{to date every professor} ought to be
\begin{equation}
\label{e:ststst}
    \bigl((s\to t)\to s\to t\bigr)\to s\to t
\end{equation}
or an even larger type.  The type~\eqref{e:ststst} is precisely equal to
$\CCSHANmonad_{cr}({s\to t})$, but simply assigning $\CCSHANmonad_{cr}({s\to t})$ as
the semantic type of \CCSHANphrase{to date every professor} would be against our
preference for the reader monad morphism to be the only component of the
semantic system that knows about the type $s$.  Instead, what we would like
is to equip the transformation on types taking each $\alpha$ to
$\bigl((s\to\alpha)\to s\to t\bigr)\to s\to t$ with a composition method
for transformed values.

\pagebreak
One tentative idea for synthesizing such a composition method is to replace
the call-by-value translation described in~\S\ref{s:translating-monads}
with a call-by-name translation, such as the Algol translation mentioned
earlier~\shortciteA[\S3.1.2]{benton-appsem}.\footnote{Another possible
translation is the standard (``Haskell'') call-by-name
one~(\citeNP[\S8]{wadler-comprehending};
\shortciteANP[\S3.1.1]{benton-appsem}).  It produces strictly larger types
than the Algol call-by-name translation,
for instance $s\to\bigl(s\to(s\to\alpha)\to s\to t\bigr)\to s\to t$.}  For
every type $\tau$, this translation recursively defines a type
$\CCSHANxlated{\tau}$:
\begin{alignat}{2}
\label{e:xlate'-type}
\CCSHANxlated{\iota}          &= \CCSHANmonad_1\iota, &\qquad
\CCSHANxlated{\tau_1\to\tau_2}&= \CCSHANxlated{\tau_1}\to\CCSHANxlated{\tau_2},
\end{alignat}
where $\iota$ is any base type.  For every term $e\CCSHANtyin\tau$, this
translation recursively defines a term $\CCSHANxlatef{e}\CCSHANtyin\CCSHANxlated{\tau}$:
\begin{equation}
\label{e:xlate'-term}
\begin{alignedat}{2}
\CCSHANxlatef{e_1(e_2)}       &= \CCSHANxlatef{e_1}\bigl(\CCSHANxlatef{e_2}\bigr), &\qquad
\CCSHANxlatef{x}              &= x, \\
\CCSHANxlatef{\CCSHANfun x:\tau.{e}}&= \CCSHANfun x:\CCSHANxlated{\tau}.{\CCSHANxlatef{e}}, &\qquad
\CCSHANxlatef{!}              &= \CCSHANunit_1(!).
\end{alignedat}
\end{equation}
Applying this translation to a monad $(\CCSHANmonad_0,\CCSHANunit_0,\CCSHANbind_0)$ gives the
types
\begin{subequations}
\label{e:xlate'-monad}
\begin{align}
\CCSHANxlatef{\CCSHANunit_0}        &: \CCSHANmonad_1\alpha \to \CCSHANxlated{\CCSHANmonad_0\alpha}, \\
\CCSHANxlatef{\CCSHANbind_0}        &: \CCSHANxlated{\CCSHANmonad_0\alpha} \to
    \bigl(\CCSHANmonad_1\alpha \to \CCSHANxlated{\CCSHANmonad_0\beta }\bigr)
    \to \CCSHANxlated{\CCSHANmonad_0\beta}.
\end{align}
\end{subequations}
If we let $(\CCSHANmonad_0,\CCSHANunit_0,\CCSHANbind_0)$ be the
continuation monad~\eqref{e:cont-monad} and
$(\CCSHANmonad_1,\CCSHANunit_1,\CCSHANbind_1)$ be the reader
monad~\eqref{e:reader-monad}, then the type $t$ transformed is
\begin{equation*}
\CCSHANxlated{\CCSHANmonad_0t} = (\CCSHANmonad_1t\to \CCSHANmonad_1t)\to
\CCSHANmonad_1t = \bigl((s\to t)\to s\to t\bigr)\to s\to t,
\end{equation*}
as desired.  However, unless $(\CCSHANmonad_1,\CCSHANunit_1,\CCSHANbind_1)$
is the identity monad, the types in~\eqref{e:xlate'-monad} do not match the
definition of monads in~\eqref{e:monad}.  In other words, though our
call-by-name translation does give the type transform we want as well as a
composition method in some sense, its output is not a monad morphism.

\section{Conclusion}

In this paper, I used monads to characterize the similarity between several
semantic accounts---for interrogatives, focus, intensionality, variable
binding, and quantification.\footnote{Other phenomena that may fall under
the monadic umbrella include presuppositions (the error monad) and dynamic
semantics (the state monad).}  In each case, the same monadic composition
rules and mostly the same lexicon were specialized to a different monad.
The monad primitives $\CCSHANunit$ and $\CCSHANbind$ recur in semantics with striking
frequency.

It remains to be seen whether monads would provide the appropriate
conceptual encapsulation for a semantic theory with broader coverage.  In
particular, for both natural and programming language semantics, combining
monads---or perhaps monad-like objects---remains an open issue that
promises additional insight.

\textbf{Acknowledgments\quad}Thanks to Stuart
Shieber, Dylan Thurston, Chris Barker, and the anonymous referees for
helpful discussions and comments.

\bibliographystyle{chicago}
\bibliography{ccshan.bib}

\end{document}